\documentclass[conference]{IEEEtran}
\usepackage{cite}
\usepackage{amsmath,amssymb,amsfonts}
\usepackage{algorithmic}
\usepackage{graphicx}
\usepackage{textcomp}
\usepackage{xcolor}
\usepackage{subfig} 
\usepackage[hyphens]{url}
\usepackage{fancyhdr}
\usepackage[bookmarks=true,breaklinks=true,letterpaper=true,colorlinks,citecolor=blue,linkcolor=blue,urlcolor=blue]{hyperref}
\usepackage{amsmath}

\fancypagestyle{firstpage}{
  \fancyhf{}
  
  \fancyhead[C]{\vspace{10pt}\normalsize{Xilinx/AMD -- Project year 2022--2024 }\\\vspace{-25pt}} 
  \fancyfoot[C]{\thepage}
}

\title{Weight Block Sparsity: Training, Compilation, and AI Engine Accelerators  } 
\author{
  \IEEEauthorblockN{\IEEEauthorblockA{P. D'\!Alberto}}
  \and
  \IEEEauthorblockN{\IEEEauthorblockA{T. Jeong      }}
  \and
  \IEEEauthorblockN{\IEEEauthorblockA{A. Jain       }}
  \and
  \IEEEauthorblockN{\IEEEauthorblockA{S. Manjunath  }}
  \and
  \IEEEauthorblockN{\IEEEauthorblockA{M. Sarmah     }}
  \and
  \IEEEauthorblockN{\IEEEauthorblockA{S. Hsu        }}
  \and
  \IEEEauthorblockN{\IEEEauthorblockA{Y. Raparti    }}
  \and
  \IEEEauthorblockN{\IEEEauthorblockA{N. Pipralia   }}
}

\begin{document}
\maketitle
\thispagestyle{firstpage}
\pagestyle{plain}


\begin{abstract}

Nowadays, increasingly larger Deep Neural Networks (DNNs) are being
developed, trained, and utilized. These networks require significant
computational resources, putting a strain on both advanced and limited
devices. Our solution is to implement {\em weight block sparsity},
which is a structured sparsity that is friendly to hardware. By
zeroing certain sections of the convolution and fully connected layers
parameters of pre-trained DNN models, we can efficiently speed up the
DNN's inference process. This results in a smaller memory footprint,
faster communication, and fewer operations.

Our work presents a vertical system that allows for the training of
convolution and matrix multiplication weights to exploit 8x8 block
sparsity on a single GPU within a reasonable amount of time. Compilers
recognize this sparsity and use it for both data compaction and
computation splitting into threads. Blocks like these take full
advantage of both spatial and temporal locality, paving the way for
fast vector operations and memory reuse. By using this system on a
Resnet50 model, we were able to reduce the weight by half with minimal
accuracy loss, resulting in a two-times faster inference speed. We
will present performance estimates using accurate and complete code
generation for AIE2 configuration sets (AMD Versal FPGAs) with
Resnet50, Inception V3, and VGG16 to demonstrate the necessary synergy
between hardware overlay designs and software stacks for compiling and
executing machine learning applications.
\end{abstract}

\section{Introduction}

To reduce computational cost of a large-scale DNNs, pruning and
quantization have been widely used \cite{Jeong2020}, which reduce the
complexity of neural networks by removing output filters all together,
part of them, or simplifying type and number of the basic operations.
Weight {\em Block sparsity} is another approach to speedup of DNN's
inference, orthogonal to pruning or quantization, and it can be used
together with them.

\begin{figure}[ht]
\begin{center}
  \subfloat[][Dense weights]{\includegraphics[width=0.6\linewidth]{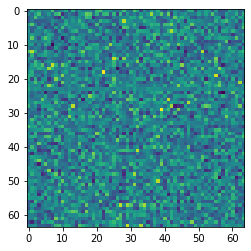}}
\vfil
  \subfloat[][Block-sparse weights]{\includegraphics[width=0.6\linewidth]{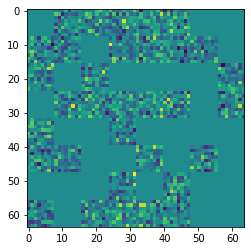}}
\end{center}
\caption{Visualization of dense and block-sparse weight matrix, zero
  blocks are green without variations} 
\label{fig:visual}
\end{figure}

Sparsity means that value of a subset of weights is exactly zero.  If
a value of a weight is zero, then the linear operation associated with
the value can be skipped, since any value times zero equals zero
(obviously).  Therefore, the computational cost of sparse linear
operations should be only proportional to the number of non-zero
weights \cite{Gray2017}.  The main problem of operations using weights
with arbitrary sparsity (i.e., also known as unstructured) is that
they cannot be efficiently implemented on contemporary hardware, such
as CPUs, GPUs, and NPUs. That is, they cannot use effectively vector
operations. However, highly optimized block-sparse operations, with
block sizes as small as $8\times 8$, can run efficiently on AIE2
processors, {\em an AI hardware accelerator}.  Figure \ref{fig:visual}
gives a visual comparison of block-sparse weights with $8\times 8$
block and dense weights.

\begin{figure}[ht]
\centerline{\includegraphics{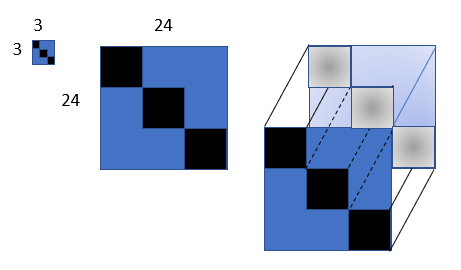}}
\caption{Example of block sparsity $\Gamma({\bar{\bf W}},8\times 8)$,
  ${\bar{\bf W}}$, and ${\bf W}$}.
\label{fig:spar_1}
\end{figure}

Consider a convolution ${\bf Y} = {\bf W}*{\bf X} + b$: The output
${\bf Y}$ is a tensor of size $M\times N \times O$, the input ${\bf
  X}$ is of size $L\times J \times I$, and the kernel ${\bf W}$ is of
size $O \times H \times K \times I$ (the bias $b$ is a vector of size
$O$). We abstract the kernel dimensions one level higher into a matrix
${\bar {\bf W}}_{i,j} = {\bf W}_{i,*,*,j}$ of size $O \times I$ where
we hide the kernel dimensions $H \times K$ (think them as going inside
the paper, Figure \ref{fig:spar_1}). We describe the granularity of a
block by a pair $b_o\times b_i \in [ 8\times8, 4 \times 16, 16 \times
  4]$. This particular granularity set is based on AIE2, but our
approach can be extended to any computational units. Of course, the
choice of granularity makes a difference in execution time in more
than one way. The {\bf block sparsity} is a process of introducing
zeros into rectangular blocks in ${\bar{\bf W}}$ and we identify such
a process as
\begin{equation}
  \begin{split}
    \Gamma({\bar{\bf W}},b_o\times b_i)_{\ell,\iota} & = 0 \\
    & \mbox{ s.t. } {\bf W}_{i,*,*,j} =0 \mbox{ with }  \\
    & i \in [\ell*b_o,(\ell+1)*b_o-1], \\
    & j \in [\iota*b_i,(\iota+1)*b_i-1] \\
  \end{split}
\end{equation}
The block $\Gamma({\bar{\bf W}},4\times 16)_{1,2}$ is zero means ${\bf
  W}_{i,*,*,j}$ is zero with $i \in [1*4,2*4-1]$ and $j \in
[2*16,3*16-1]$. In practice, $\Gamma$ is a bit map or a mask (i.e., 0
zeros, 1 unaffected but it could be used as weighted mask with values
in $[0,1]$).  We can express block sparsity percentage as the relative
number of zeros in $\Gamma$ and in particular the number of zeros in
every row of $\Gamma$, where $N$ is the number of columns in $\Gamma$
and for every row of $\Gamma$:
\begin{equation}
  {\cal S} = \frac{1}{N}( N-\sum_i^N\Gamma({\bar{\bf
      W}},O\times I)_{o,i} )
\end{equation}

In Figure \ref{fig:spar_1}, we show an example where we zero one block
per row and the block granularity is $8\times8$.

The main advantages of block sparsity are:

\begin{enumerate}
  \item The non-zero blocks are dense in the inner dimensions (input
    channels) and we can tune the granularity. We exploit spatial
    locality by reading in one cycle 4 elements and we read a multiple
    contiguous in memory. This is spatial locality and it must be
    exploited by long-row memories and long-line caches. In turn, it
    is highly used by {\em vectorizable} functional units that consume
    and produce multiple results in a single clock.
  \item The block are dense in output dimension so the same weight
    volume can be reused for multiple outputs; that is, better
    throughput and possibly latency.
   \item The zeros blocks are easily skipped: 50\% sparsity results in
     50\% fewer operations with the identical throughput and latency.
  \item The computation is naturally split by block of filters,
    exploiting (output) parallelism, and each sub-computation has a
    known and countable number of operations per row and per block.
\end{enumerate}

Block Sparsity would be beneficial for any hardware architecture with
vectorizable instructions (spatial locality and parallelism) and
multiple computation units (thread parallelism). We exploit block
sparsity in terms of training neural network, its compilation, and
implementation on a certain hardware. The main contributions of this
work are:

\begin{enumerate}
\item We proposed Weight Block Sparsity method on a pre-trained neural
  network, which is hardware-friendly structured method, and can be
  applied any CPU, GPU, and NPU (Neural processing Unit) and reduce
  the memory footprint of a deep neural network and accelerate its
  inference.

\item Also, we proposed the compiler and code generation for the
  neural network with Block Sparsity.  Our Compiler is based on
  parameterized representation of block sparsity, allows the
  capability to split tensors and computations accordingly to a
  parameterized representation of the architecture.

\item Lastly, we could estimate the time execution for operations of a
  neural network including DDR to mentile communications and graph
  computation.  Sparsity reduces computation and communication time,
  and is captured using the time estimation.
\end{enumerate}

\section{Related Works}

The modern reference of sparsity in deep learning is by Hoefler et al
\cite{Hoefler2021}, which is an entire spectrum of sparsity and
is not limited to block sparsity. The original sparsity comes from
applications, for example \cite{Saad2003,Davis2006,Saad2020}. Data
structures and algorithms are tailored to exploit sparsity. Often, the
data structure is tailored to the sparsity in order to compact data,
different sparsity structures require different algorithms, and
achieve different performance. Block sparsity in this context is for
matrices (i.e., often diagonal matrices) where a block is stored
contiguously in memory. Siswanto in \cite{siswanto2021} introduces
block sparsity in convolutional network generalizing for matrices
(i.e., fully connected layers) but the application to convolution
layers is different than ours (the author does pruning of the kernel
by height or width). Our block sparsity describes logically a block
but not in memory layout, in most framework the {\em input channel} is
the innermost dimensions. The zeros are shuffled in, as for matrices,
we shall compact the blocks.

The works by Numenta in \cite{Ahmad2019,Hunter2022} go into two
directions: block sparsity of the convolution kernels and of {\em
  activation} tensors; that is, input and output. For the kernel, they
can choose any granularity and every filter can be sparsified so that
a jig-saw puzzle can be created: each filter is block sparse with
different patterns so that, across filters, they can further compact the
data structure. The sparsification of the activation tensor does not have a pattern; it is called unstructured
sparsity. The structured sparsity can be achieved only by a carefully
process that must be part of the network training. It is not clear
from the literature how the training is carried on.

Nvidia implemented 50\% fine-grained sparsity on Ampere architecture
\cite{Pool2020}.  Nvidia A100 supports convolution and matrix multiply
operations in 2:4 pattern, 2 values are zero out of each contiguous
value of 4.  The sparsity network achieved 1.3x to 1.5x in speedup.
The same functionality is available in AMD GPUs and (future) AIE2s
instruction sets. This sparsity is appealing because of its hardware
support, deterministic space footprint, and computation
reduction. However, it imposes very stringent constraints on the type
of sparsity not always allowing a comparable accuracy.

The difficulty of generating code for complex architectures can be
described and solved in different ways. There are recent attempts of
introducing Software/Hardware solution for spatial processors
\cite{Huang2021CoSASB,Russo2023MemoryAwareDA,Cai2023InterlayerSS}.
However, their major attentions are given only to matrix
multiplication and GPUs \cite{Gray2017GPUKF}
\cite{li2023popsparse}. In our case, we are focusing on the so called
static sparsity or weight sparsity.

\section{Block Sparsity in a Matrix}

Block sparsity is an intuitive concept but it can be
misunderstood. Take a matrix multiplication in Equation \ref{eq:mat}
\begin{equation}
  \label{eq:mat}
  \begin{pmatrix}
    C_{0} & C_{1} \\
    C_{2} & C_{3}  
  \end{pmatrix} = 
  \begin{pmatrix}
    A_{0} & A_{1} \\
    A_{2} & A_{3} \\ 
  \end{pmatrix}\\  \begin{pmatrix}
    0   & B_{1} \\
    B_{2} & 0 \\ 
  \end{pmatrix}\\
\end{equation}
This is the computation 
{\small \begin{equation} 
    C_{0} = A_{1}B_{2}; \; 
    C_{1} = A_{0} B_{1}; \; 
    C_{2} = A_{3} B_{2}; \; 
    C_{3} = A_{2} B_{1}
\end{equation}}
and in general with proper $\gamma_i$ (i.e., a mask)
\begin{equation}
  C_{i} = \sum_{k=0}^1 A_{i+ k} \big(\gamma_{2*k+i} B_{2*k+i}\big)
\end{equation}
Where the matrix $B$ is constant, diagonal, and each submatrix $B_{2}$
and $B_{1}$ can split further down and may have even smaller zero
blocks. For data structure in the sparse computation field, we can use
{\em compress block row} (CBR) or {\em column format} (CBC) or
generalization of the {\em coordinate format} (COO). There are
standard matrix sparse-matrix multiplication interfaces and algorithms
for CPU and GPUs using this data format (where only one operand is
sparse or both) \cite{rocSPARSE,cuSPARSE}. In the introduction
section, we describe block sparsity applied to convolutions. We store
a compact $\Gamma$ for inference, this could be literally a bitmap or
an block row/column index. For the effective computation by the AIE
kernel, we deploy a relative row/col index that will apply to dense
operations as well. The figurative sparsity is now expressed in a
computation where the savings are clear.

This is briefly our overall process and contributions.  We explore
training techniques using PyTorch and Keras in Section
\ref{sec:training}.  We compute a mask $\Gamma$ of zeros/ones per
layer by zeroing the more likely blocks (using a Norm). Then we train
the model till convergence or accuracy are achieved. We take the
sparse model and we quantize to 8-bit integer computations with a
quantizer which is based on round-to-nearest quantization (Equation
\ref{eq:quant}). The final model is an intermediate representation.
We have a custom compiler that takes the intermediate representation
and an abstraction of a connected set of AIE2 (i.e., Section
\ref{sec:HW} and \ref{sec:compilers}).  The compiler computes the
maximum sub-volume computation per core. By heuristics and following a
schedule, it computes a memory allocation in Memtile (i.e.,
intermediate scratch pad Section \ref{sec:memory}) for inputs,
outputs, and weights . It formats, compresses, and organizes the
weights exploiting spatial distribution to Memtiles and cores. We
generate all the explicit communications between DDR ($L_3$), Memtile
($L_2$), and cores ($L_1$). These are Gather and Scatter instructions
with a complete and parametric estimate of their execution time by
bandwidth constraints and number of channels. The calculation is based
on the sub-volume sizes per core, the computation throughput and with
a clear specification of what is executed in parallel. The data
movement codes and AIE core codes for simple convolutions were
simulated and run in hardware for simple layers (i.e., Section
\ref{sec:code}). Thus, we can achieve consistent execution time
estimates per layer and of the entire network with an accuracy closer
to a simulation (realistic although optimistic Section
\ref{sec:timeestimate}).  We will show estimates for three CNN models
and eight different AIE designs; see Section \ref{sec:experiments}.
To our knowledge, we are the first in applying sparsity on AIEs
systematically. The ability to provide different algorithms and easy
to obtain estimates for very different configurations, it will allow
to explore optimizations like sub-graph depth-wise tiling we could not
have otherwise.

In our context, convolution is our main computation and CNN are
networks we can train reasonably. This is because of legacy, we want
to speed up the FPGA work-horses, convolutions provide more
difficulties than GEMMs (padding, strides, and overlap), have usually
biases with different precision requirements (weight 8bits and bias
32), routinely they can deploy different scaling factors per output
channels, and GEMM is transformed into a $1 \times 1$ convolution
immediately. The other way around is possible with proper activation
preparation; in fact, a convolution can be represented as a GEMM but
the activation tensors explode in space by a factor of the kernel size
and the preparation is more suitable for CPUs than FPGAs.

\subsection{Block-Sparse Matrix-Matrix Multiplication}
Consider $\Gamma$ and $\Omega$ two matrices
in $\{0,1\}^{N\times N}$.
\begin{equation}
  C = (\Gamma A) * (\Omega B)^t
\end{equation}
More precisely, consider non-zero blocks of size $k\times k$ so that
\begin{equation}
  C_{i*N+j} = \sum_k ( \gamma_{i*N+k} A_{i*N+k} ) (\dot{\omega}_{j*N+k} \dot{B}_{j*N+k})
\end{equation}

\begin{figure}[ht]
\begin{center}
  \subfloat[][]{\includegraphics[width=0.8\linewidth]{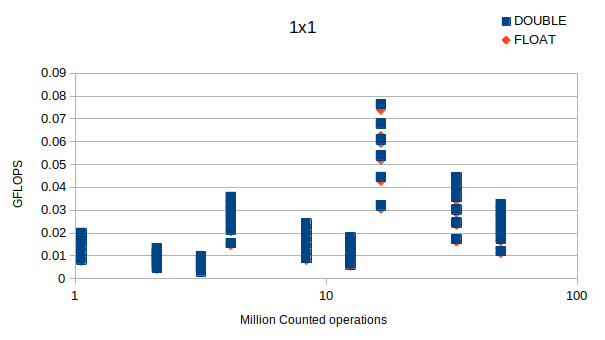}}
\vfil
  \subfloat[][]{\includegraphics[width=0.8\linewidth]{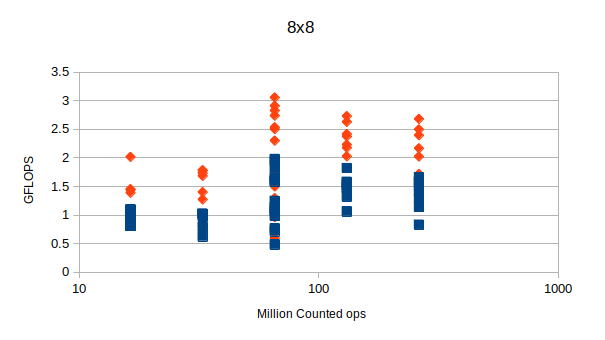}}
\end{center}
\caption{Block 1x1 and 8x8
  performance} 
\label{fig:block}
\end{figure}

Thanks to the sparsity and if we store only non-zeros, then
$\gamma_{i*N+k}$ and $\dot{\omega}_{j*N+k}$ are contiguous. There will
be a meaningful product to compute if and only if $\gamma_{i*N+k} =1$
and $\dot{\omega}_{j*N+k} =1$.  We merge-sort these vectors.  See how
the sparse sparse matrix multiplication using {\em Coordinate Block
  Structure} (COO) is applied in Figure \ref{fig:block}.
We provide software to reproduce this \cite{PaoloG2020}.

In the case of achieving a fixed sparsity (i.e., density) of 50\% for
a square matrix of size $N$ and we choose the block size $k \times
k$. The larger $k$ is, the smaller the overhead will be.  The relative
performance of the $k^3$ multiplication is better as $k$ get larger
because spatial, temporal locality, and further optimized code for a
constant/parameterized such as $k$.

In Figure \ref{fig:block}, we present two scatter plots: on the
abscissa the effective multiplication-and-addition number, on the
ordinate the performance in GFLOPS, when the sparse matrix with dense
block is $1\times 1$ (above) and $8\times8$ (below). Given the same
task, we deploy more threads and thus the vertical effect (AMD 16
cores Threadripper).  With the same number of effective operations,
the block permits and exploits higher GFLOPS per effective operation
(Float is 2x faster than Double precision and this can be emphasized
further \cite{Gray2017GPUKF,li2023popsparse} and
\cite{pmlr-v119-kurtz20a}, for AIE2 we can do 256 multiply accumulate
per cycle using 8 bit operands, 128 per 16 bits, 64 per 32bits).

\section{Block Sparsity: Training and Quantization}
\label{sec:training}
The block sparsity we plan to deploy is not naturally
recurring in Convolutional Neural Networks.  
Training is required and achievable practically as fine tuning of trained networks. 

As reminder, a convolution has a weight tensor in four dimension: $W
\in R^{c_{out}\times h \times k \times c_{in}}$. In the hyperplane of
the $h$ and $k$, we can simplify the weight as $\dot{W} \in R^{c_{out}
  \times c_{in}}$ and block sparsity can be simply described by a mask
$\Gamma\dot{W}$. Although, we speak of a $8\times 8$ of non-zeros,
this is in practice a $8\times h\times k\times 8$ block. For the
matrix multiply $h=k=1$, there is no difference from the unstructured
sparsity. We explain the training process.

\subsection{Searching Optimum Sparsity ratio}

We target convolutions first and without quantization. We take a model
and we create a copy where we enhance the convolution with a
(non-trainable) $\Gamma$. $\Gamma$ is the sparsity ratio and mask.
This experiment is conducted using Keras\cite{chollet2015keras} and
the code is available \cite{PaoloK2020}. A convolution will have three parameters
(saving the model into a different format).  The forward computation
is modified so that the weights used for convolution are $\Gamma
{W}$. We assume the backward computation (i.e., gradient) is done
automatically from the forward definition. There is no need to change
the bias. For example, we take Resnet50 from the Keras application
repository, we start with a $\Gamma=1$, and we trained one epoch using
Imagenet dataset \cite{deng2009imagenet}.  The goal is to choose
$\Gamma$ in such a way we achieve the required sparsity and the
minimum loss in accuracy.  We start from $W_{0}$, the current optimal
weight, then choose values in $W_{0}$ to make zeros using different
approaches such as incremental, Fisher measure, Hessian, diagonal
Hessian, and custom penalty losses.

\subsection{Sparsity Ratio as Incremental}

For example, for every convolution, $n_i =\sum \gamma_i$, which is the
number of blocks, reduce this number to $\frac{n_i}{2}$. For each
epoch (say every two training epochs), we consider the current
non-set-yet mask elements $\sum (1-\gamma_i) = k < \frac{n_i}{2}$. We
compute our importance measure for all in ascending order. This time,
we zero the first $\min(\frac{5}{100}k, 1)$. We keep this process
until we reach 50\% sparsity. At each iteration at least one block
will be set to zero. We trace a solution path as much as geodesic as
possible.

\subsection{Sparsity Ratio as Trainable as Optimization Problem}

If we want to make $\Gamma$ part of the optimization process as
trainable variable we could introduce a penalty function into the loss
${\bf \ell}(x,w) + \lambda(w)$. First let us introduce an
approximation for the $\max(x)$, so when in this section you will read
$\max, \min$, this is a log sum exponential, which is continuous,
derivable everywhere, and convex:

\begin{equation}
  \max(x) = LSE(x,\alpha) = \frac{1}{\alpha}\log \sum e^{x_i*\alpha} 
\end{equation}

With $T$ we represent the number of non-zero block in $\Gamma$

\begin{flalign}
  \lambda=  &  -(\max(\Gamma)- \min(\Gamma))  &&\\\nonumber
  & +\beta*L2(\Gamma-T) + \iota*L1(\Gamma)  &&
\end{flalign}

This is a simplified loss so that we minimize the value of $\Gamma$ but
also try to maximize the difference of the elements.

\begin{flalign}
  \lambda=  & \max(-\Gamma,0) + \max(\Gamma -1, 0)  -(\max(\Gamma)- \min(\Gamma)) &&\\\nonumber
            & + \beta*L2(\Gamma-T) + \iota*L1(\Gamma) &&
\end{flalign}

This last penalty function represents our attempt to state that the
$\gamma_ i$ should be positive and in the interval $[0,1]$ and in a
such a way that we maximize the distance of the elements between 0s
and 1s.
\begin{flalign}
  \lambda=  & \max(-\Gamma,0) + \max(\Gamma -1, 0)  -\frac{\min(\Gamma)}{\max(\Gamma)} &&\\\nonumber
            &+ \beta*L_{2}(\Gamma-T) + \iota*L_{1}(\Gamma)   
\end{flalign}

\subsection{Hessian and Fisher Information}
If we have $N$ parameters/weights, the Hessian $H \in R^{N\times N}$
has quite the space complexity (consider even small models could have
million parameters). When we are already close to the optimal solution
or we are trying to move from the optimal solution without using
information from the gradient, the Hessian provides the most
information close by an already established solution point. There are
also ways to compute the Hessian and the effects of the Hessian by
using Fisher information
\cite{yao2020adahessian,abs-2101-08940,zandonati2022fit}. This will
reduce to the computation of the gradient of the loss function.

\subsection{Diagonal Hessian}
We applied a Fisher measure and computed $\nabla^2{\bf \ell}$, that is
computed just the diagonal of the Hessian. Again, we use the $L_2$
over the normalized weight and went through the process of
training. The elements of the diagonal are not representative in
general, but they are a good approximation of the contribution of a
single weight.  The Fisher and $\nabla^2{\bf \ell}$ did not provide
any main advantages. But this information is very useful in
quantization and optimizations within the same field and application.

\subsection{Predetermined Sparsity ratio and Full Training Ahead}

Take a convolution with $\Gamma = 1$ and weights $W$. Once again for each
$\gamma_i$, this will be representative of a block $W_{i} \in R^{8
  \times h \times w \times 8} \sim R^{8\times 8}$. We can choose the
$W_{i}$ using a measure of importance:

\begin{itemize}
  \item $L_{2}(W_{i}) = \sqrt{\sum_k w_k^2}$ with $w_{k} \in W_{i}$,
  \item $L_{1}(W_{i}) = \sum_k |w_{k}|$ as above,
  \item Variance $\sigma^2 = \frac{1}{64}\sum_{k} (w_{k} -\mu)^2$ with
    $\mu = \frac{1}{64}\sum w_{k}$, with $w_{k} \in W_{i} $ or $\frac{1}{N}\sum
    w_{k}$, with $w_{k} \in W$. 
\end{itemize}

We can then sort them in ascending order. We set the first half to
zero.  Then we start re-training. We do this for the entire network or
for one convolution at a time.

In Table \ref{tab_acc}, we show the results by using one-time mask and
full training: VGG-16, ResNet-50, Inceptionv3 on ImageNet20 (20
classes) and ImageNet1k (1000 classes).  We use three samples per
class for the validation accuracy for ImageNet1k data set; instead, we
use 50 samples per class for ImageNet20. Fine-tuning sparse networks
on the original ImageNet data-set \cite{deng2009imagenet} is
expensive. To reduce the training time, we chose 20 classes (from the
original 1000 classes) with the least number of images per class in
the training data-set and this choice will affect the accuracy because
there are fewer samples for re-training.

\begin{table}[ht]
\caption{Accuracies of the sparsity models}
\label{tab_acc}
\begin{center} 
\scalebox{0.9}
{
\begin{tabular}{|l|c|c|c|c|c|}
\hline
\rule[-1ex]{0pt}{3.5ex}  Model & Dataset & Baseline  & \multicolumn{3}{c|}{Sparsity}\\
\rule[-1ex]{0pt}{3.5ex}  {} & {} & Acc.(\%) & block & ratio (\%) & Acc.(\%)    \\\hline\hline
\rule[-1ex]{0pt}{3.5ex}  Inception-v3 & ImageNet1k & 77.2 & 8x8 & 50 & 75.5  \\\hline
\rule[-1ex]{0pt}{3.5ex}  ResNet-50 & ImageNet1k & 76.7 & 8x8 & 50 & 74.6  \\\hline
\rule[-1ex]{0pt}{3.5ex}  VGG-16    & ImageNet1k & 70.6 & 8x8 & 50 & 69.7  \\\hline \hline
\rule[-1ex]{0pt}{3.5ex}  ResNet-50 & ImageNet20 & 96.1 & 8x8 & 25 & 95.1  \\\hline
\rule[-1ex]{0pt}{3.5ex}  ResNet-50 & ImageNet20 & 96.1 & 8x8 & 50 & 92.0  \\\hline
\rule[-1ex]{0pt}{3.5ex}  ResNet-50 & ImageNet20 & 96.1 & 8x8 & 75 & 87.1  \\\hline
\rule[-1ex]{0pt}{3.5ex}  ResNet-50 & ImageNet20 & 96.1 & 1x1 & 25 & 96.0  \\\hline
\rule[-1ex]{0pt}{3.5ex}  ResNet-50 & ImageNet20 & 96.1 & 1x1 & 50 & 95.6  \\\hline
\rule[-1ex]{0pt}{3.5ex}  ResNet-50 & ImageNet20 & 96.1 & 1x1 & 75 & 93.5  \\\hline
\rule[-1ex]{0pt}{3.5ex}  VGG-16    & ImageNet20 & 92.0 & 8x8 & 50 & 89.6  \\\hline
\rule[-1ex]{0pt}{3.5ex}  VGG-16    & ImageNet20 & 92.0 & 1x1 & 50 & 92.3  \\\hline
\rule[-1ex]{0pt}{3.5ex}  VGG-16    & ImageNet20 & 92.0 & 1x1 & 75 & 91.7  \\\hline
\end{tabular}\vspace{-20pt}
}
\end{center}
\end{table}

Classification accuracy on ImageNet1k drops by only 1-2\% after
applying 50\% sparsity with a $8\times 8$ block (this is without any
quantization). We experiment with different block shapes such as
$16\times 4$ and $4\times 16$ on ResNet-50, but the accuracy is
slightly worse. The different shape has the same volume and either
more parallel vector operations or longer vector computations.  This
last process based on PyTorch has been the most accurate while the
other approaches had a drop of 7 points in accuracy at least. It is
not clear if it is a methodology weakness or just bad execution. It is
important to us to present all the tools attempted.

Fine-grained sparsity ($1\times 1$ block or unstructured) does not
sacrifice any accuracy (i.e., almost any). This is not equivalent to 2
over 4 (or 4 over 8) sparsity now available in GPUs.

\section{Compiler and its Code generation}
\label{sec:compilers}
We take a PyTorch/Keras model and quantize it before creating an intermediate representation.
Consider a weight $W$. The linear operation can be written as $y=Wx$, and the quantized one is $y=W_{q}x$. The quantization function is defined as:

\begin{equation}
\label{eq:quant}
  W_{q} = \Delta * Round(W/\Delta) \mbox{ with }  \Delta = \frac{max(|W|)}{2^{N-1}}
\end{equation}

Our intermediate representation is a graph where each node is an
operation that reads tensors and writes one tensor.  A convolution has
one quantized input INT8 with a position where the fraction starts
(power of two scale). It computes a tensor using the same layout and
with a proper scale.  The weights and bias are properties of the
convolutions. They are tailored and laid out at compile time, they are
$C_{OUT}\times h \times w \times C_{IN}$

The compiler in this work can create the parameterized representation
of block sparsity, and has the capability to split tensors and
computations accordingly to a parameterized representation of the
architecture. Our Hardware abstraction is a Python class, describing a
variety of systems. All weights are statically prepared into DDR and
we move them explicitly towards the inner levels. Inputs and outputs
have designated space in DDR. DDR can and it will be used for tensors
spills.  The memory allocation to Memtile is basically coloring
algorithms and some heuristics. In this architecture, we do not allow
{\em streaming} of neither data nor weights (because they share space
in Memtile and input and output have different consumption/production
rates).

\subsection{Hardware Abstraction}
\label{sec:HW}
\begin{figure}[ht]
\begin{center}
  \subfloat{\includegraphics[width=0.7\linewidth]{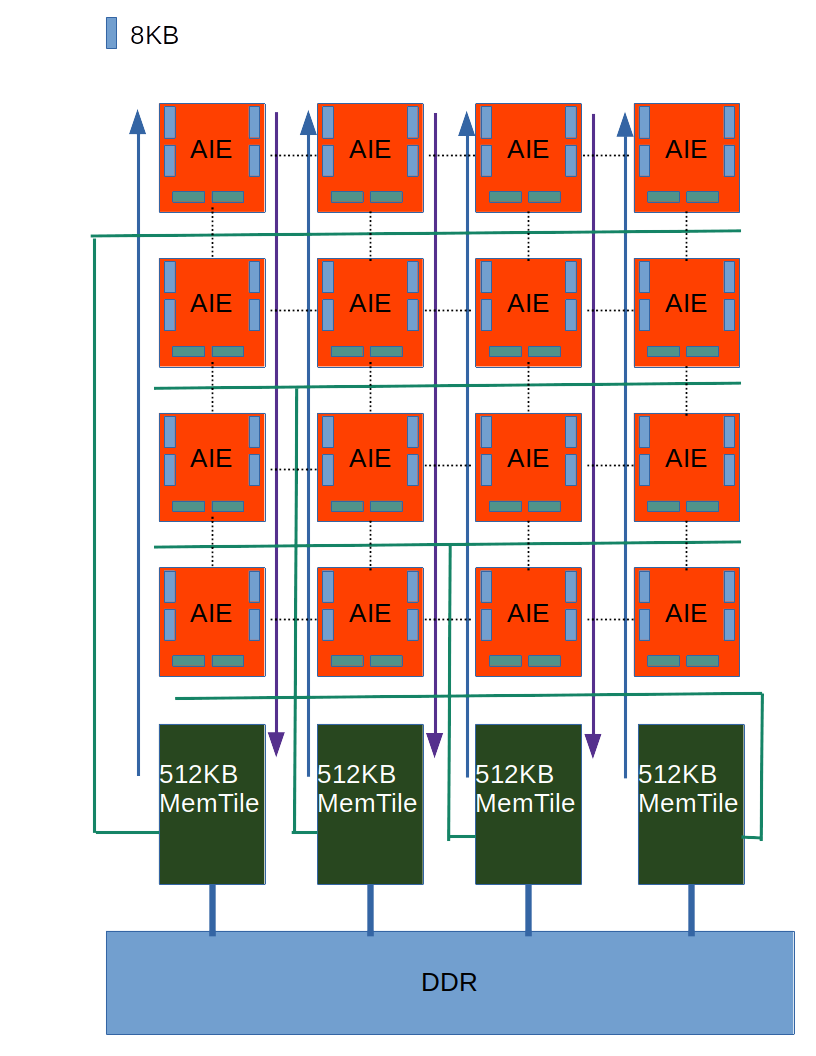}}
\end{center}
\caption{4x4 AIE representation} 
\label{fig:aie}
\end{figure}

As shown in Figure \ref{fig:aie} as an example, we work with a mesh of
4x4 tensor cores connected by 4 horizontal and 4 vertical
interconnections.  We present estimates for square 2x2, .. $i\times i$
.. 8x8 and rectangular shapes are in the works ($4\times 1$, $4 \times
2$, and $8\times 2$ into a $8 \times 8$ with 2 Memtiles per
column). Each core has 8 banks memories for a total 64 KB. About six
banks are used as input/output/weight buffers and two banks are used
as temporary space for kernels. Each core can request and send data to
its direct neighbors (if aware of connection and control but this
utility is not used here). Double buffering using ping/pong is used
for inputs and outputs.

There are four Memtiles: each 512 KB and each is connected to one
columns and its direct neighbor column, or it is connected to a row
and its neighbor. The total amount of space is 2 MB. Memtile is a
circular buffer to exploit more flexible allocation. Note a $2 \times
2$ architecture will have one Memtile per column and a total of two
Memtiles (1 MB).

A Memtile can broadcast data per column or per row; it is a design
choice. We can dedicate one Memtile for weights, one for activations,
or we can share it. In this work, we present results for shared
Memtiles. To maximize the computation parallelism, every core will
write data per column into Memtile.

\subsection{Subvolumes, Data Compression, and Data Movements}
The computation is split by Memtile and thus by column (cores
columns). The output tensor is computed and split evenly by
width. Thus one Memtile will store one partial tensor by width, each
core will compute different output channels, and the computation
streams the output tensor by rows and using ping/pong double
buffering. We prioritize to reuse weights in core. The cores set is a
cohort and we always choose symmetric computations. 

If we have the inputs, output, and weights in Memtile, 
the minimum computation is
one output channel and one row (i.e, by height). If this is not
possible, we try to reduce the size of the width (e.g., shaping the
tensor in Memtile by using DDR spills) and we can manage to split the
input channels and to split the weights accordingly and prepare for
accumulation. We call W-Split the distribution of tensor by columns
in the Tensor mesh. We can $C_{OUT}$-split, this requires the partial transfer
of weights.  We can $C_{IN}$-split when we need to split by input channel,
this is the last resort because it is also the most expensive
accumulation of the outputs. $C_{IN}$-split can be implemented as a graph
optimization by splitting the convolution into two and then use an
element wise operation to combine the results, which can be done
recursively.

The subvolume describes the smallest shape of the weights that we need
to manage and the largest computation in the core. We compress the
weight accordingly. Any data movement will always be a multiple of the
subvolume, a multiple of $8\times 8$,  and it is a single {\em load}. 
Such a compressed data will have the
same properties whether it is sparse or dense.

\subsection{Schedule and Memory Allocation}
\label{sec:memory}
During the scheduling of each layer, we evaluate what tensors can fit
in Memtile. Here, activation and weight tensors share the space.  At
each step, the memory allocation will check if we can allocate
inputs, weights, and outputs. If we cannot, we evict all tensors
into DDR and then split the time of the computation.

At the end of this stage, every tensor will have an address in Memtile
or DDR (or both). If there are only DDR addresses, the compiler will
take the basic layer computation and, by heuristics, will split the
computation and the output tensor by width, output channel, height,
and input channel (no necessarily in this order). The heuristics have
a single objective to find the largest problem fitting the (each)
memory level. We deploy a recursive approach of tiling.  Formally,
$\dot{\sum}$ is a parallel loop and a W-split can be written as
follows:

\begin{equation}
  Y =  Conv(X,W) = \dot{\sum}_w
  Conv(X_{w},W)
\end{equation}

The split is a function of the footprint. Before and after each
convolution, there will be an explicit data movement as option. At
this stage each input, output, and weights have addresses associated
with each sub-computation. Then the code generation of each
$Conv(X_{w},W)$ is independent and recursive as needed. This
is a tree-like structure. If the convolution has strides or a large kernel, each
sub-convolution has overlap data; however, the sub-convolution has
defined addresses and data movements. For a W-split such as this, we
are computing the output by rows and the weights are reused (read
once).

\subsection{Code Generation }
\label{sec:code}
\begin{figure}[ht]
\begin{center}
  \subfloat{\includegraphics[width=1.5\linewidth,angle=90]{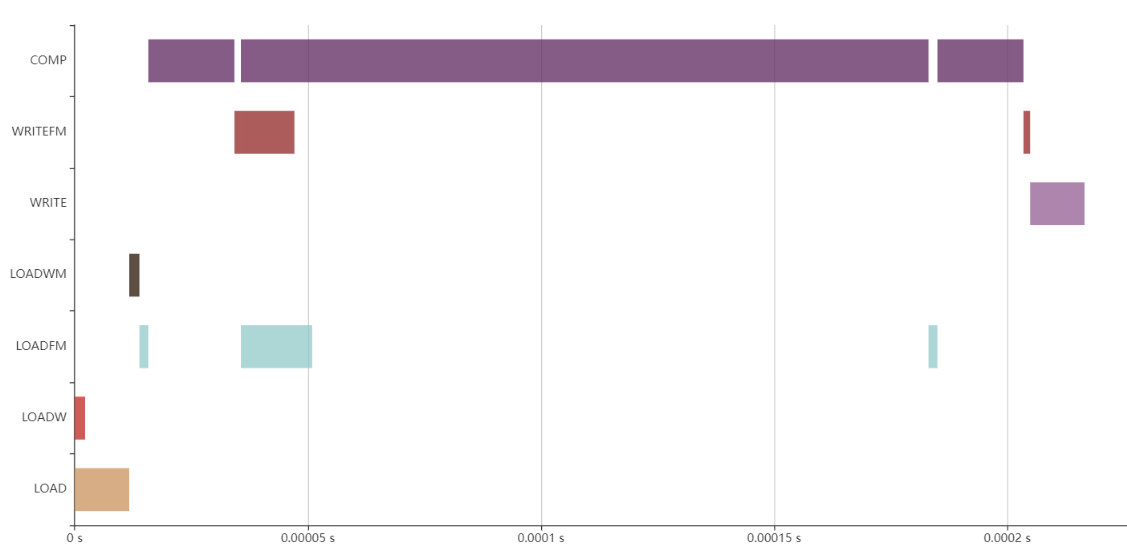}}
\end{center}
\caption{Resnet single convolution
  with padding for 4x4: LOAD activation from DDR to
  Memtile, LOADW weights from DDR to Memtile, LOADFM activation from
  Memtile to Tensor cores, LOADWM weights from Memtile to Tensor cores, WRITE
  from Memtile to DDR, WRITEFM from Tensor Cores to Memtile, COMP Computation in this case a convolution.} 
\label{fig:singleconvestimate}
\end{figure}

The compiler creates a list of operations. These operations become
smaller and smaller and then can be executed from Memtile to
Memtile. There is a further decomposition using only the Tensor cores and it
is completely determined by the subvolume. Here, we show how we
generate code at this level and estimate time as in Figure
\ref{fig:singleconvestimate}.  This is the computation of a
convolution with top/bottom padding by height:

\begin{equation}
  \label{eq:convpadding}
  Y_{height} =   Conv(X_{h=0}) \dot{+} \dot{\sum}_{h=1}^9
  Conv(X_{h}) \dot{+} Conv(X_{h=10})
\end{equation}

An important feature of the current system is the concept of {\bf
  iteration} between Memtile and core.  Using locks and chaining the
locks, we write a single instruction from the prospective of a single
core (as a SIMD instruction) and driving all cores at once for
multiple iterations $\dot{\sum}_{h=1}^i Conv(X_{w})$ in Equation \ref{eq:convpadding}. The ASM-like code follows:

{\footnotesize
\begin{verbatim}
  LOADFM Lock k_0 Memtile addr core addr iter i
  CONV iteration      i
  WRITEFM Lock k_1 Memtile addr core addr iter i
\end{verbatim}
} 

There is an implicit lock (say \verb2k_x2) that is used for the pong
and the system cycles between locks (\verb2k_x2 and \verb2k_02).
These three operations execute a number of iterations {\em i} and,
using a ping/pong, they will load different slices of data and compute
different slices of data. 

Equation \ref{eq:convpadding} is encoded as follows: {
\footnotesize
\begin{verbatim}
  ## Head top pad < 50 us First comp block
  LOADFM Lock k_0 Memtile addr_0 core addr iter 1
  CONV iteration 1
  WRITEFM Lock k_1 Memtile addr_1 core addr iter 1
  ## Body iteration > 50 us < 150 us
  ## k_0 -> k_2 -> k_4 Lock Chain
  LOADFM Lock k_2 Memtile addr_2 core addr iter 9
  CONV iteration 7
  WRITEFM Lock k_3 Memtile addr_3 core addr iter 9
  ## tail bottom pad > 150 us Last computation block
  LOADFM Lock k_4 Memtile addr_4 core addr iter 1
  CONV iteration 1
  WRITEFM Lock k_5 Memtile addr_5 core addr iter 1
\end{verbatim}
 } 
 
 We present in Figure \ref{fig:singleconvestimate} the execution
estimate of this code. At this stage, we have all the information. Per
layer, the code generation is a two pass process. First, we generate
code for the all loads/stores. Second we combine them into chains with
dependency, logically correct and as fast as possible.

We could estimate the time execution without a full code
generation. When we annotate time information to a load, we have
assurance that the load is a complete description of the DMA
communication between multiple memories and as complex as the
architecture. Actually, this is literally translated to a binary
executable that perform the data movement.

\subsection{Time Estimation}
\label{sec:timeestimate}
We explain how we capture the execution time and visualize it as in
Figure \ref{fig:singleconvestimate}. We start by the time estimates
for DDR to Memtile communications. We have two communication types:
activations and weights. Per Memtile there are two dedicated channels.
\begin{itemize}
 \item If we share activations and weights in the same Memtile, we can
   use one channel for activations and one for weights. Thus the loads
   from DDR to Memtile (LOAD and LOADW) are parallel with a bandwidth
   of 4 GBps. Writes from Memtile to DDR (WRITE) can use both channels
   (8 GBps).

 \item If activations and weights go to different Memtiles (for
   example weights to Memtiles '0' and '3' and activations to '1' and
   '2'), each load is parallel and 8 GBps. Writes are identical.
\end{itemize}
   
The Memtile connections with AIE cores are different. We assume a few
channels with 4 GBps bandwidth. One Memtile can broadcast inputs to a
cores column. These communications are for activations (LOADFM). One
Memtile can broadcast to rows of cores, these are for weights
(LOADWM). We assume that the column and row communications are
parallel.

Every communication with iteration {\em one} is synchronous and
sequential.  The load, convolution, and store is executed one after
the other and every core is independent.  For synchronization and for
bookkeeping, we assume that Weights communications from Memtiles to
cores are synchronous and halting (LOADWM).

Every communication with iteration larger than one, we assume that
load (LOADFM), computation (COMP), and store (WRITEFM) are executed in
parallel and the overall execution time is the maximum of the
estimated time multiplied by the number of iterations.

We estimate the execution time of a subvolume (COMP) by the number of
operations divided by the maximum number of operations per cycle which
is in our scenario is $4\times 8 \times 8 = 256 $ operations per cycle
and 1 GHz frequency. Sparsity reduces computation and communication
time.

We do not account the wait and synchronization which are necessary to
reprogram the fabric. These are very expensive running on for a few
milliseconds.

\subsection{Sparse Convolution example}

\begin{figure}[ht]
\begin{center}
  \subfloat{\includegraphics[width=1.5\linewidth, angle=90]{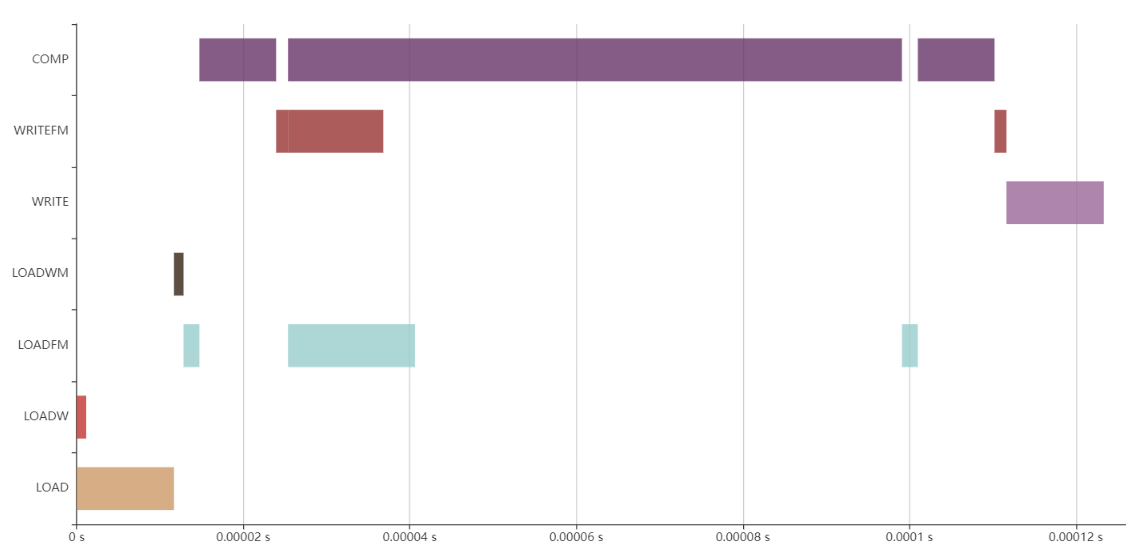}}
\end{center}
\caption{Resnet single convolution
  with padding and sparsity for 4x4 Tensor Cores} 
\label{fig:singleconvestimate2}
\end{figure}

We present the time estimate for a convolution with padding, dense,
and with 50\% sparsity, see Figure \ref{fig:singleconvestimate} and
\ref{fig:singleconvestimate2}.

\begin{figure}[ht]
\begin{center}
  \subfloat{\includegraphics[width=1.5\linewidth, angle=90]{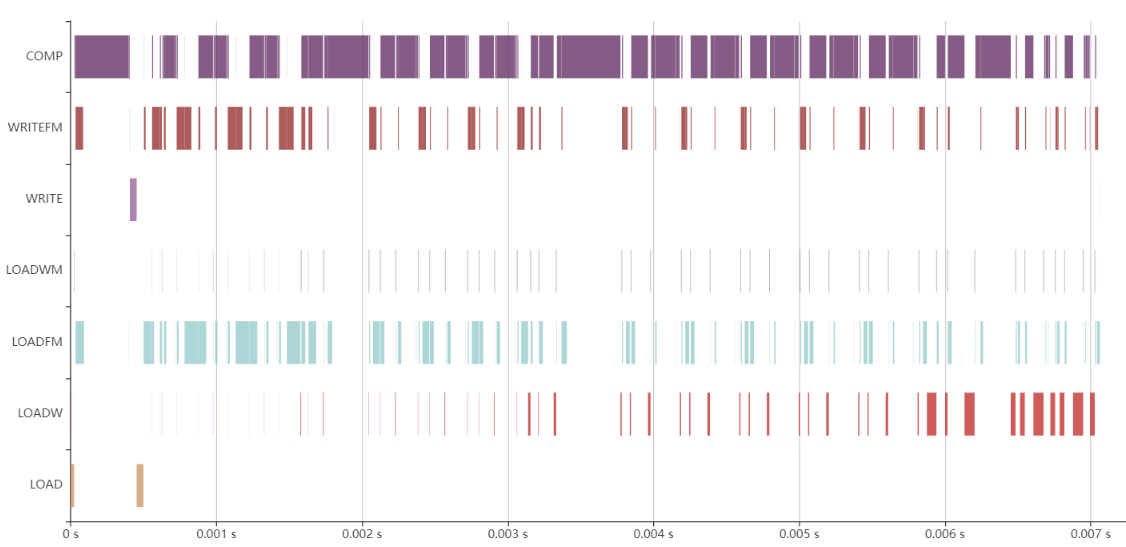}}
\end{center}
\caption{Resnet50 for 4x4 Tensor Cores with 50\%
  sparse weights} 
\label{fig:estimate-sparse}
\end{figure}

For these convolutions, the computation dominates the execution
time. Sparsity cuts the execution time by half: from 200 $\mu s$ to
130 $\mu s$. On one hand, there are convolutions that realize up to
$2\times$ performance; on the other, there are convolutions that are
dominated by the reading or writing. In the latter case, sparsity
helps in space saving and probably DDR tensors spilling. In principle,
we could relax sparsity requirements for those convolutions that are
communication bound (and restart training). In Figure
\ref{fig:estimate-sparse}, we provide the time estimates for Resnet50
using sparsity 50\% and using $4\times 4$ AIE array.

\section{Depth-wise Tiling}

Take the subgraph schedule $L_{0}, L_{1}, \dots L_{j}$, for every $i$
$L_{i}$ is a layer that produces a single tensor $T_i$. Depth-wise
tiling is the idea of tiling the output tensor $T_{j}$ into tiles
$u_{m}$, then cut the computation accordingly to find the minimum
corresponding input shape and size of $T_{-1}$ and $T_{0}$, even
overlapping, to carry the computation to $u_{m}$. Let us define $u_j =
1\times1$, where we neglect the {\em channels}, as a sub-tensor of
layer $L_{j}$; that is, a sub-vector of $T_{j}$. A projection $\P$ is
a function taking $u$ and projects the input sub-tensor needed to
compute $u$.  In reverse order, starting from $L_{j}$ compute
$\P(L_{j},u_{j})$ and propagate the projection. When a layer $L_{m}$
and tensor $T_{m}$ feeds two (or more) layers $L_{x}$ and
$L_{y}$. Then $u_{m} = \max ( \P(L_{x},u_{x}),\P(L_{y},u_{y}))$ when
$u_{x}$ and $u_{y}$ are defined completely and {\em max} mean the
largest. We carry the propagation of $\P(L_{m},u_{m})$ and we have
$T_{-1} =\P(L_{0}, u_{0}) =\P(L_{0} \P(L_{1},u_{1}))=\P(L_{0}, \dots
\P(L_{l}, u_{l})$.  At the end of the propagation, imagine this is a
generalized convolution with kernel size $k = \P(L_{0}, u_{0})$. Now
if we repeat the process with $\dot{u_{j}} = 2\times 2$ ... $k+s =
\P(L_{0}, \dot{u_{0}})$. Thus we have an estimate of the generalized
stride.

Now we can split the computation by selecting a non overlapping
decomposition of the output tensor: say using M tiles each of size
$u_o$. The input is split into M tiles of size $(u_o-1)s+k$ and we
will have an overlap of size $k-1$. We can choose a tiling in such a
way there is zero DDR communication in between $L_{0}$ and $L_{j}$.
With input overlap, thus with extra reads, we have extra
computations. However, we can reduce drastically the DDR
communications. We use the term generalized convolution but it is not
a convolution and it is only applied to a graph computation (with
element wise operations and transpose convolutions, we do not think it
is applicable to height and width softmax or layer norms).

\subsection{Real-time analysis and Zero DDR Communication}
Assume, we have an output tile size $u_{o}$. We repeat the projection
process using the schedule. This time, we keep information of the
active tensors at any time in the schedule. Projections and active
tensors are sufficient to represent the maximum active space
requirement (not including input and output). If we want zero DDR
spill for inputs and outputs, we choose a output tile for which the
maximum active space is smaller than Memtile. Say we have M
tiles. With an architecture for which computation is free (see
\cite{Hong1981IOCT,BilardiPD00}), we write $T_j$ once, we read $T_0$,
and we reread $(M-1)(k-1)$. This is better than $\sum_{l=0}^j T_l$
especially for large $j$s.

A compiler can target easily a tiling with zero DDR communications
(for layer output tensors). In practice, zero DDR communications is
not our final goal.

\subsection{Iteration and Memory allocation}
Take the schedule $L_{0}$ \dots $L_{j}$, let us apply the live tensor
analysis and tensor projections for a specific tile size $u_{o}$ and $M$
tiles. We add two custom layers as follows
\[ I_{b}(M), L_{0}, L_{1}, ... L_{n}, I_{e}(M) \]

The layer $I_{b}(M)$ is an iteration layer or input boundary that takes
the input tensor $T_{-1}$ and carves out a sub-tensor shape and size
$(u_{o}-1)s+k$. The layer $I_{e}(M)$ is an iteration layer or output
boundary that takes the input tensor $u_{o}$ and copies it to a tensor
of shape and size $T_{j}$. For the purpose of the memory allocation, we
have a legal schedule that describes basically the property of the
computation.  We can do memory allocation.

\subsection{Code Generation}
A computation is basically defined by its layer and by the addresses
in memory of its input and output tensors. The code generation unrolls
the iterations as it was a loop and $I_{b}$ and $I_{e}$ generate the
proper sub tensor addresses and specifying the number of iterations.

The code generation will be for the schedule:
\[ I_{b}(0), L_{0}, L_{1}, ... L_{n}, I_{e}(0), 
\dots I_{b}(M-1), L_{0}, L_{1}, ... L_{n}, I_{e}(M-1) \]
With a little of book keeping we can write the time estimates as a
time series and compare performance.

\subsection{Complementary: depth-wise tiling and sparsity}
The depth-wise tiling has the tendency to increase the computation
number and decrease the DDR loads and stores. As fomulated, this
tiling may read multiple times the weights. Sparsity reduce the number
of computations and the weights communications. Together have the
opportunity to work better.

\begin{table}[htb]
  \caption{Execution Time estimates}
  \label{tab_perf}
\begin{center} 
\begin{tabular}{|l|l|l|l|l|}
  \hline
  AIE2 & Model  & Dense sec      & Sparse sec      \\ \hline\hline
  2x2   & Resnet & 2.492347e-02  & 1.582626e-02 \\ \hline
  3x3   &  & 1.269543e-02  & 8.661490e-03 \\ \hline
  4x4   &  &  1.077318e-02 & 7.064918e-03 \\ \hline
  5x5   &  &  failed       & 4.303485e-03 \\ \hline
  6x6   &  &  5.712521e-03 & 4.490127e-03 \\ \hline
  7x7   &  &  4.205991e-03 & 3.212234e-03 \\ \hline
  8x8   &  &  6.376768e-03 & 4.602027e-03 \\ \hline \hline
  2x2   & IncV3  & 4.283837e-02  & 2.440544e-02 \\ \hline
  3x3   &   & 2.386600e-02  & 1.422390e-02 \\ \hline
  4x4   &   &  1.740967e-02 & 1.012540e-02 \\ \hline
  5x5   &   &  9.690552e-03 & failed       \\ \hline
  6x6   &   &  1.063962e-02 & 6.439692e-03 \\ \hline
  7x7   &   &  8.727651e-03 & failed       \\ \hline
  8x8   &   &  9.093276e-03 & 5.666152e-03 \\ \hline \hline
  2x2   & VGG16  & 4.476212e-02  & 2.608593e-02 \\ \hline
  3x3   &   &  2.53343e-02  & 1.002015e-02 \\ \hline
  4x4   &   &  1.371000e-02 & 8.852128e-03 \\ \hline
  5x5   &   &  failed       & 4.336479e-03 \\ \hline
  6x6   &   &  failed       & 5.770197e-03 \\ \hline
  7x7   &   &  7.455440e-03 & 5.288551e-03 \\ \hline
  8x8   &   &  9.203393e-03 & 6.502333e-03 \\ \hline \hline
          
\end{tabular}
\end{center}
\end{table}

\section{Performance of Sparsity}
\label{sec:experiments}
In Table \ref{tab_perf}, we present the performance of neural networks
with sparsity, where the sparsity is applied to all the convolutions
(except the first one because there are only three channels and
sparsity requires at least eight) for Resnet 50, Inception V3, and
VGG16.  We estimate the total execution time for three networks and
seven configurations in Table \ref{tab_perf}. We report also the case
where the compiler fails to generate code. Notice that the
configuration $8\times 8$ AIEs is never beneficial. A full
investigation is necessary.

Corner cases are represented as failure in Table \ref{tab_perf}. Some
cases is because of inability to break the weight tensor evenly.
Sometime is for incorrect data management especially for prime
numbers. These are all issues that we will address as the technology
mature. Please, note that VGG16 using 8x8 configuration is slower than
7x7 (by using sparse).  For a symmetric computation too small
sub-volume computations make the computation overall more inefficient
and requiring more iterations for the same amount of data
transfers. This is a case where more HW does not improve performance,
which is interesting and relevant.

\subsection{Depth-Wise Tiling for VGG16 3x3}
We present results for VGG16 because of simplicity and each layer
tensor do not fit the three Memtile (of a $3\times 3$ system). We can
apply the same approach to Resnet and inception. The generalized
convolution idea is applicable.

\begin{figure}[ht]
\begin{center}
  \subfloat{\includegraphics[width=1.8\linewidth, angle=90]{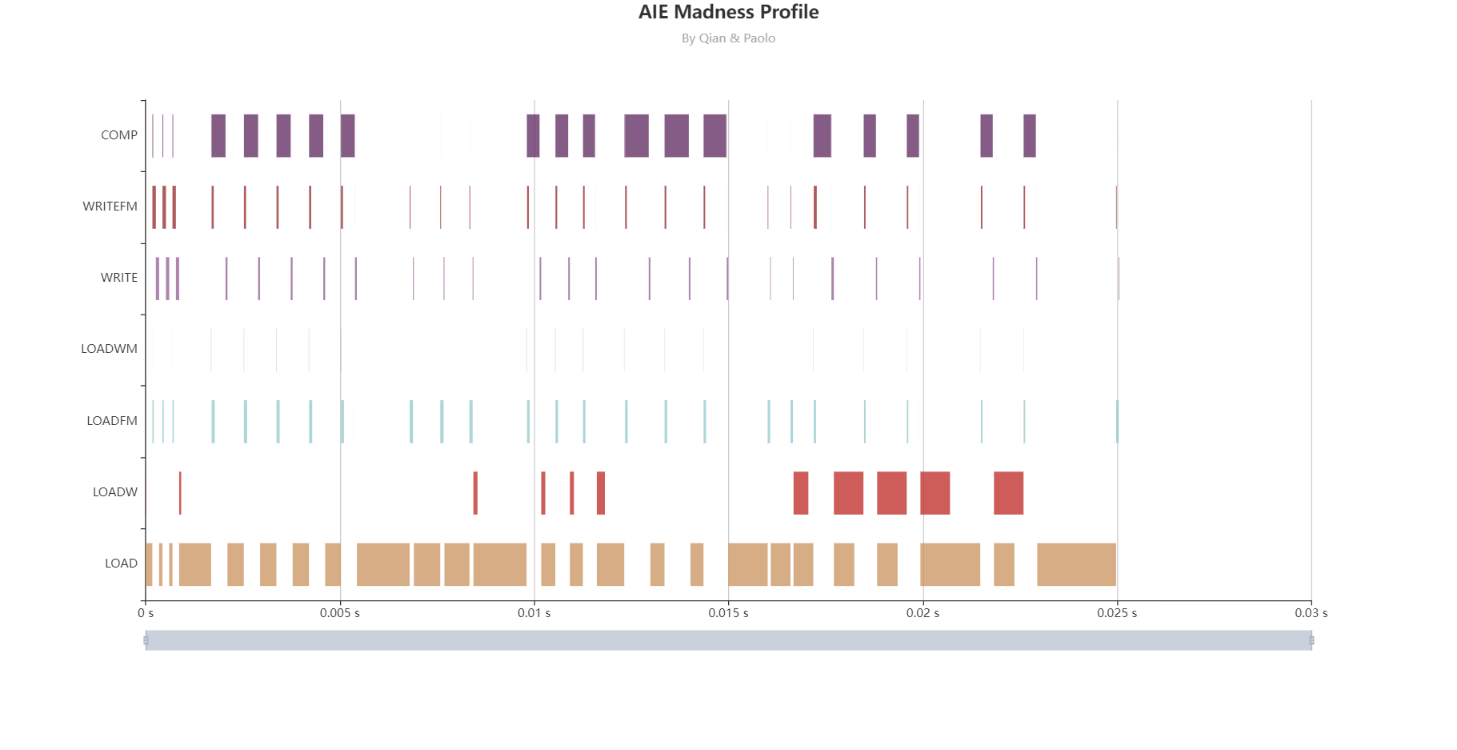}}
\end{center}
\caption{VGG16 3x3 DDR only} 
\label{vggddronly}
\end{figure}

We take VGG and we instruct the DDR to be 16 times slower (4GBs/16)
highlighting the need of fewer DDR communications. We take only the
part that requires DDR spills for each layer. 
In Figure \ref{vggddronly}, we show the performance for VGG16 using DDR spills for each layer: 0.025s total time.

\begin{figure}[ht]
\begin{center}
  \subfloat{\includegraphics[width=1.8\linewidth, angle=90]{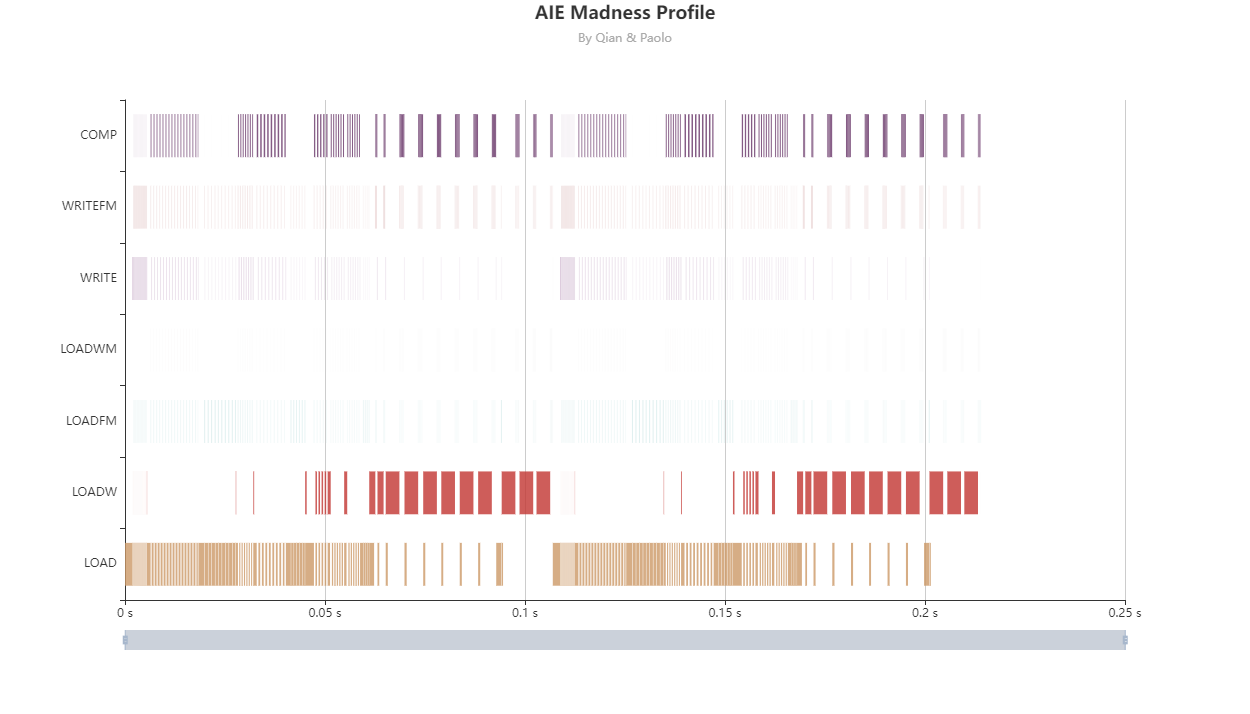}}
\end{center}
\caption{VGG16 3x3 DDR with 2 tiles} 
\label{vgg2tiles}
\end{figure}

\begin{figure}[ht]
\begin{center}
  \subfloat{\includegraphics[width=1.8\linewidth, angle=90]{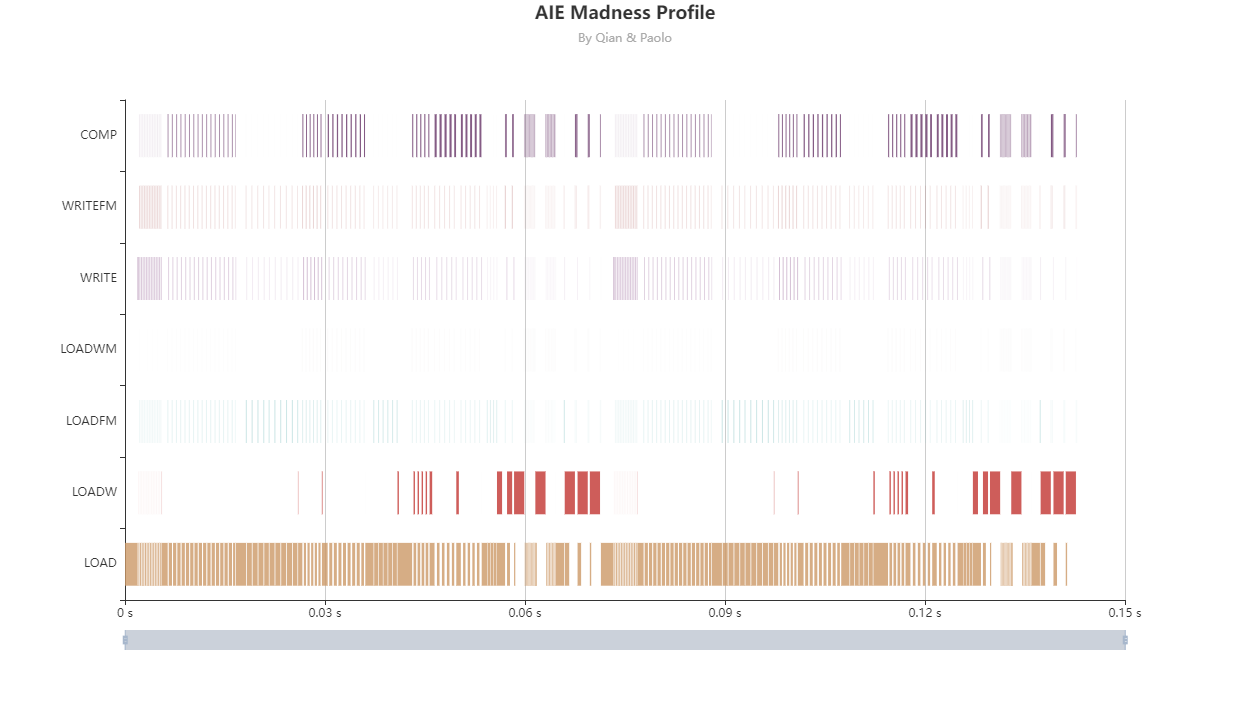}}
\end{center}
\caption{VGG16 3x3 DDR with 2 tiles and sparse} 
\label{vgg2tilessparse}
\end{figure}

We apply depth-wise tiling using two tiles and we achieve better
performance at 0.022s. see Figure \ref{vgg2tiles}.  Sparsity by itself
without any tiling can achieve only 0.021s.  Sparsity and tiling
improves even further and we achieve 0.014s, see Figure
\ref{vgg2tilessparse}. We can appreciate the reduction of activation
DDR communications thanks to depth-wise tiling and the reduction of
computation and weights communication by sparsity.

\section{Conclusions}
This is a multifaceted problem and we present a complete solution from
training techniques, compilers, code generation, Hardware definition, and
time estimations. It is a vertical software system, more complex than
just a prototype, and it is used for the validation and comparison of
different Hardware designs. A few convolutions have been validated in
simulation and in hardware.

This could be seen as a quantization and sparsification problem. For
example, how we can reduce the footprint of a CNN network. There are
post training techniques that are targeting quantization and
unstructured sparsity \cite{frantar2023gptq}. We need to be more
aggressive and training for it.  Our sparsity is not really a property
of the model, software can describe it, and the hardware can take
advantage; however, we do not need specific hardware support at
instruction level. To our knowledge we are the first applying sparsity
to Tensor Cores such as AIE2 overlays systematically.

We demonstrated Sparsity can accelerate the computing in Matrix
multiplication and Convolution in neural networks.  Matrix
multiplication is appealing for the application in LLM and application
in GPUs. Convolutions is far richer in complexity and it is the
work-horse for FPGAs based products and systolic array
systems/computations.



\bibliographystyle{IEEEtranS}
\bibliography{refs}

\end{document}